\documentclass[10pt,a4paper]{article}
\usepackage{authblk}
\usepackage{setspace}
\usepackage{amsmath}
\usepackage{graphicx}
\usepackage{subfig}
\usepackage{float}
\usepackage{booktabs}
\usepackage{numprint}
\usepackage{url}
\usepackage[square,sort,comma,numbers]{natbib}

\usepackage[labelfont=bf,labelsep=colon]{caption}
\title{Deep learning for prediction of population health costs}
\author[1]{Philipp Drewe-Boss*}
\author[2]{Dirk Enders}
\author[2]{Jochen Walker}
\author[1]{Uwe Ohler}

\affil[1]{ Berlin Institute for Medical Systems Biology, Max Delbr\"{u}ck Center for Molecular Medicine in the Helmholtz Association, Robert-R\"{o}ssle-Strasse 10, 13125 Berlin, Germany}
\affil[2]{Institute for Applied Health Research (InGef), Spittelmarkt 12, 10117 Berlin, Germany}

\begin{document}

\maketitle

\begin{abstract}
Accurate prediction of healthcare costs is important for optimally managing health costs. However, methods leveraging the medical richness from data such as health insurance claims or electronic health records are missing.
Here, we developed a deep neural network to predict future cost from health insurance claims records.
We applied the deep network and a ridge regression model to a sample of 1.4 million German insurants to predict total one-year health care costs. Both methods were compared to Morbi-RSA models with various performance measures and were also used to predict patients with a change in costs and to identify relevant codes for this prediction.
We showed that the neural network outperformed the ridge regression as well as all Morbi-RSA models for cost prediction. Further, the neural network was superior to ridge regression in predicting patients with cost change and identified more specific codes.
In summary, we showed that our deep neural network can leverage the full complexity of the patient records and outperforms standard approaches. We suggest that the better performance is due to the ability to incorporate complex interactions in the model and that the model might also be used for predicting other health phenotypes.
\end{abstract}

\section{Introduction}

\begin{figure}[h]
	\begin{center}
		\includegraphics[width=1.0\textwidth]{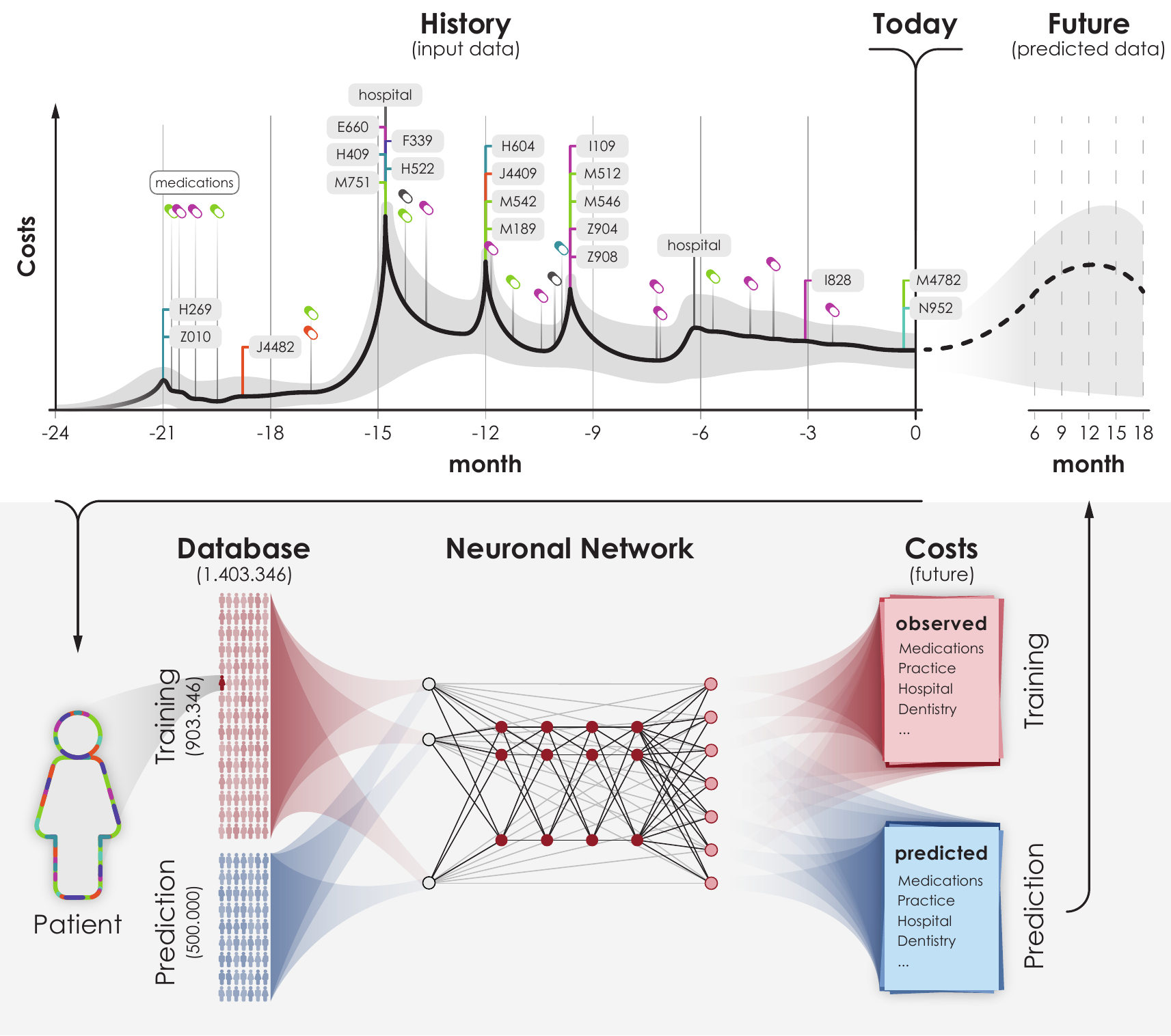}
	\end{center}
	\caption{\textbf{Graphical abstract}}
	\label{Fig:ProjectOverview}
\end{figure}

Health care expenditures are one of the biggest expenses in Germany and optimally managing these cost has great economical importance.
Therefore, methods for accurate patient-level prediction of future health care cost are needed to provide the basis for decision making.
As medical costs reflect the development of health over time, and health in turn is influenced by many factors such as social demographics, previous medical history, environmental influences, genetics but also by random events such as accidents, predicting the future health is inherently challenging. Consequently, accurately predicting health cost is a challenging problem.
Existing work on prediction of health cost can be divided into two categories~\cite{Sushmita2015}: (1) Rule based prediction methods, in which decision rules of an algorithm to predict future costs are manually defined. The disadvantage of this approach is that it requires deep domain knowledge and that the capability of resulting models to reflect complex relations in the data is limited. (2) Supervised learning based methods (e.g. linear regression models, random forests or support vector methods) that learn to predict future cost from the data~\cite{Bertsimas2008, lahiri2014predicting, Sushmita2015, drosler2017sondergutachten, Frees2013, Duncan2016}. These methods have the advantage that they are not limited in their expressiveness as rule based methods are. However, they typically require large datasets for training.
For training of these methods, health insurance claims records are an appealing data source. They cover most of the health care expenditures of the patients and have the advantage of having sample sizes that allow fitting rich models. Additionally, they contain detailed information on patients, such as the medical history and social demographic information. 
The challenges of this data is that it is high dimensional, that there are many hidden interactions between variables, and that the data is often not normally distributed~\cite{Diehr1999}. The aforementioned supervised learning methods are believed to typically not leverage the potential of population scale data to detect complex patterns~\cite{tang2018canadian}.
Recent developments in deep learning techniques, such as novel deep neural network architectures and numerical approaches to fit the networks, promise to address some of these challenges. Deep learning has been successfully applied in the medical domain to task such as dermatologist level detection of skin cancer~\cite{esteva2017dermatologist}, prediction of various clinical outcomes from electronics health records~\cite{rajkomar2018scalable}, or the detection of diabetic retinopathy from retinal fundus photographs~\cite{gulshan2016development}, showing the potential of this technology. 

We present a novel deep neural network architecture to predict future health care cost from health insurance claims records (See Figure~\ref{Fig:ProjectOverview}). This network architecture allows to fully capture the richness of the medical data in health insurance claims and can be fitted on a standard workstation with 64GB of RAM. We compare it on health insurance claim records of $\sim1.4$ million patients from German statutory health insurances against various standard methods and show that it outperforms existing approaches. It also is better identifying patients at risk than standard linear regression approaches. Finally, we show how the parameters of the network can be interpreted and that the network uses medically relevant features for its prediction. 

\section{Methods}

\subsection{Data}
This study is based on data of the Institute for Applied Health Research Berlin (InGef) database, which contains anonymised longitudinal claims data of more than 60 German statutory health insurances. Claims data of the years 2010 to 2017 for a sample of about $1'403'346$ insurants was used, which is representative for the German population with respect to age, sex and state of residence.  Besides sociodemographic information, the database contains information on hospital stays, outpatient physician visits, drug prescriptions and remedies and aids including costs in each of the sectors. Further details of the database can be found elsewhere \cite{Andersohn2015}. An approval of an ethics committee or informed consent of the patients was not required for the conduct of this study since all patient- and provider-level information are anonymised to comply with German data protection regulations and German federal law. 
In the remainder of this manuscript we will refer to the period ranging from Q1 2010 to Q4 2015 as the observation period and to the period from Q3 2016 to Q2 2017 as the evaluation period.

\subsection{Data Representations}
The input for the machine learning algorithms was formatted in the following manner: Each numerical value was kept as a feature. Dates were coded in quarters since Q1 of 2010. Categorical values, such as International Statistical Classification Of Diseases And Related Health Problems, 10th revision, German Modification (ICD-10-GM) codes,  Anatomical Therapeutic Chemical (ATC) codes, Diagnosis Related Group (DRG) codes, German procedure classification (OPS), physician subject group key (FG) and 
schedule of fees for physician outpatient services (GOP) codes or sex, were coded using a one-hot encoding (i.e. if $n$ possible categories $k_1,\dots,k_n$ were possible, the observation of category $k_j$ was coded by a $n$-dimensional vector that was $1$ at the index $j$ and $0$ everywhere else). If multiple categories were observed in a quarter, the representing vectors were added. This coding was performed for each quarter and the resulting vectors were concatenated into a single vector of dimension 24*91'470 =2'491'470 representing the patient data in the observation period. In order to accelerate model fitting, we only considered variables that had more than $1'000$ entries, leading to a vector of dimension 24*13'876 = 333'024.  

\subsection{Model definition}
\begin{figure}[h]
\begin{center}
	\includegraphics[width=0.8\textwidth]{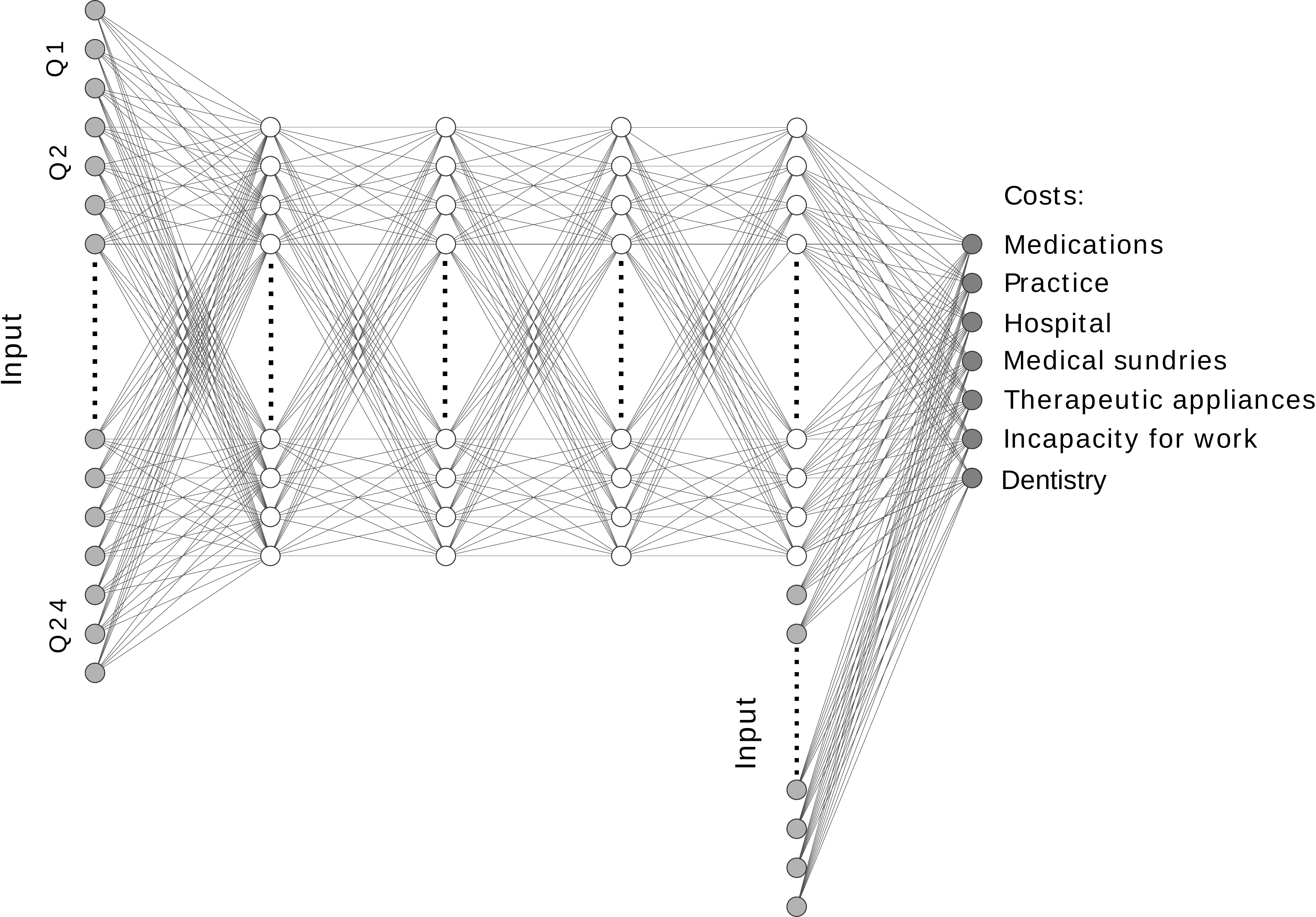}
	\caption{\textbf{Network architecture:} Shown is the architecture of the proposed deep neural network. Shown in (light grey) are the input features. Shown in (dark grey) are the target variables of the network. The (white) nodes are the internal nodes of the network.}
	\label{Fig:NetArch}
\end{center}
\end{figure}

We used a model with four hidden layers (See Figure~\ref{Fig:NetArch}). The first four layers had each $50$ neurons. In the fourth layer the original input was concatenated to the hidden vector and fed to the last layer, which had seven neurons to predict seven cost categories (Medications, practice, hospital, medical sundries, therapeutic appliances, compensation for incapacity to work and dentistry). All layers used the ReLU-activation function~\cite{nair2010rectified} and a dropout~\cite{srivastava2014dropout} rate of $0.25$ during training. 

We compared the deep learning model to three standard models. (1) The average cost per year in the previous $6$ years. (2) The costs in the last year of the observation time. (3) A ridge regression with parameter $\lambda=0.1$.
For model assessment, we predicted separately the seven (see above) different cost domains. We then summed all predicted costs except the cost to compensate for incapacity to work for model assessment, in order to make the cost comparable to costs reported for the Morbi-RSA~\cite{drosler2018,drosler2017sondergutachten}. 
Furthermore, we performed model ensembling for the ridge regression and the neural network (i.e. the network as trained five times and the predictions of all five models was averaged for the final prediction.)

\subsection{Model fitting}

For model fitting we used the first $903'346$ patients (training set). During model fitting, we minimized the $l2$-loss between the future and the predicted costs using ADAM~\cite{kingma2014adam}, which is an extension of stochastic gradient descent. Both the ridge regression and the deep learning model was trained for $25$ epochs. For training of the ridge regression a batch size of $128$ was used and for training of the neural network a batch size of $32$ was used.

\subsection{Implementation}
All models have been implemented in python and keras~\cite{chollet2015keras}.

\subsection{Evaluation Criteria}
To assess the model quality, we determined the following quality criteria: Pearson's correlation coefficient, Spearman's correlation coefficient, the mean absolute error and Cumming's Prediction Measure. The performance was evaluated on the subset of $357'239$ of the $500'000$ held out patients (test set) that where alive in the observation period and either died or were still insured on at least one day in the evaluation periods.

We further assessed how well the methods could be used to identify patients with changing costs. As this is indicating a change of health status or treatment, these patients could benefit from preventive interventions.
To this end, we divided our test set patients into three groups. Those for which the cost decreased more than $100$-fold between the last year of the observation period and the first year of the prediction period; those for which the cost increased more than $100$-fold; and the remaining patients. In order to not include patients with overall low cost in the two group (e.g to not include patients that change from 0.01 to 10.0 Euro) that have strong cost changes, we added 10 Euros to the overall cost before computing the fold change. We then computed the area under the precision-recall curve (auPRC) for identification of patients with increasing, resp. decreasing costs from all patients. To understand for which cost range the respective methods performed best, we computed the error of the prediction in dependence of the cost.

\subsection{Sensitivity analyses}
We investigated how the performance of the neural network depends on the amount of available training data. To this end, we trained the model on only $100'000$,  $200'000$,  $300'000$,  $400'000$,  $500'000$,  $600'000$,  $700'000$,  $800'000$  and $900'000$ patients. Furthermore, we investigated how the length of the observation time affects the predictive performance. Therefore, we trained the model also for each of the patient sets using the data from one to six years up to the end of the observation period. 

\subsection{Feature identification}
An important application of predictive models is to identify relevant features in the data and to understand their effect on the prediction. This allows for example to identify and quantify risk factors. A common approach in linear models is to identify the weights that have a large absolute value as they correspond to the features that have as strong impact on the prediction. For deep neural networks it has been shown that this strategy is suboptimal~\cite{sundararajan2017axiomatic} as it does not capture the interactions between features that the neural network uses. Here, we therefore used a strategy called integrated gradients~\cite{sundararajan2017axiomatic} that is more robust. We determined the average integrated gradients of all patients in the evaluation set. Furthermore, we divided the mean integrated gradient by the number that the actual feature was nonzero, to account for the fact that not all features are equally abundant. We did not show codes in the results that allow identification of health insurance companies which contributed to the study database. 

\section{Results}
To establish a baseline, we first compared the performance of all methods to predict costs. We found that the neural network was able to better predict future costs than ridge regression or the other two standard models in all considered measures as shown in Table~\ref{Tab:Perfm}. Furthermore, we found that ensembling several training runs provides an additional small improvement.

\begin{table}[]
	\begin{tabular}{llllll}\toprule
			& r                        & $\rho$ & MAPE & $r^2$    & CPM            \\\midrule
		Spendings in last year     & 0.418 & 0.551   & 2403.30  & -0.005 & 0.191 \\
		Mean of previous spendings & 0.464 & 0.547  & 2078.76 & 0.200  & 0.301 \\
		Ridge regression           & 0.514 & 0.610   & 2126.03 & 0.260  & 0.285 \\
		Neural network              & 0.524 & 0.631  & 2013.35 & 0.264  & 0.323 \\
		Ridge regression (ensemble)  & 0.517 & 0.611  & 2116.67 & 0.265  & 0.288 \\
		Neural network  (ensemble)    & \textbf{0.527} & \textbf{0.632}  & \textbf{2004.33} & \textbf{0.266}  & \textbf{0.326}\\\midrule
		Morbi-RSA model (2018)$^*$ & na & na & 2267.60 & 0.258 & 0.242 \\
		Morbi-RSA full model$^*$ & na & na & 2233.53 & 0.263 & 0.253 \\\bottomrule
	\end{tabular}
\caption{\textbf{Performance assessment}. Evaluation of methods using: Pearson's correlation (r),  Spearman's correlation ($\rho$),  mean absolute prediction error (MAPE), R squared ($r^2$) and Cumming's Prediction Measure (CPM). Performance for the Morbi-RSA models on a different data set ($^*$) where obtained from \cite{drosler2017sondergutachten, drosler2018}. Correlation values where not available (na) for these models.}
\label{Tab:Perfm}
\end{table}

\begin{figure}
		
		\begin{center}
			\includegraphics[width=\textwidth]{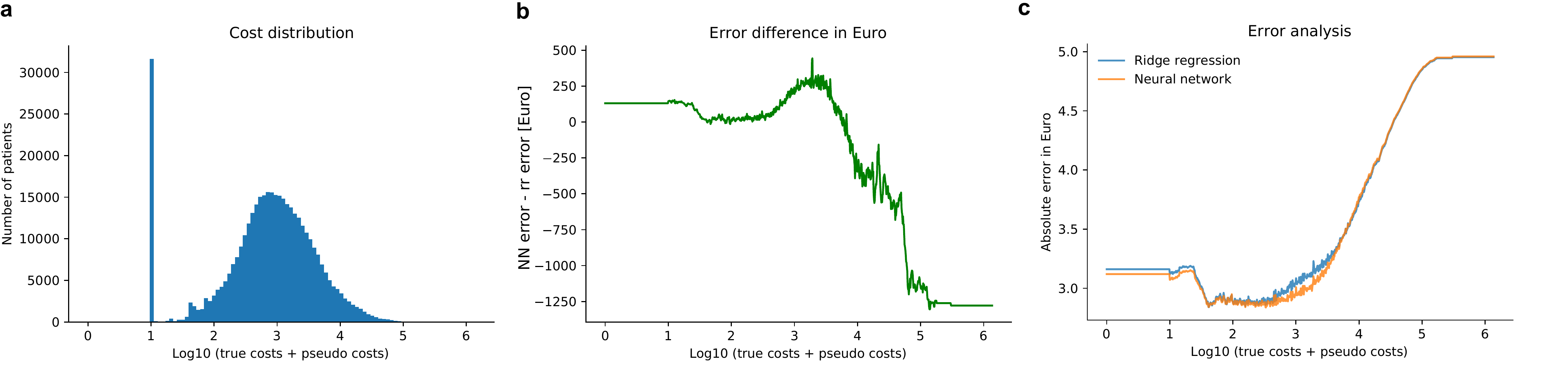}
		\end{center}
	\caption{\textbf{Error analysis:} Shown is the histogram of total cost \textbf{(a)}, the log10 absolute error based on the true cost of the ridge regression and the neural network \textbf{(b)} as well as the difference between the neural network error and the ridge regression error \textbf{(c)}}
	\label{Fig:ErrorMerged}
\end{figure}

To better understand in which cost regimes the neural network and the ridge regression performed better, we studied the average absolute error in Euros depending on the true costs. The neural network performed better for patients with total costs lower than $\sim 10'000$ Euro, whereas the ridge regression performed better for patients who were more expensive (See Figure~\ref{Fig:ErrorMerged}a and Figure~\ref{Fig:ErrorMerged}b).

\begin{figure}
	\begin{center}
		\includegraphics[width=\textwidth]{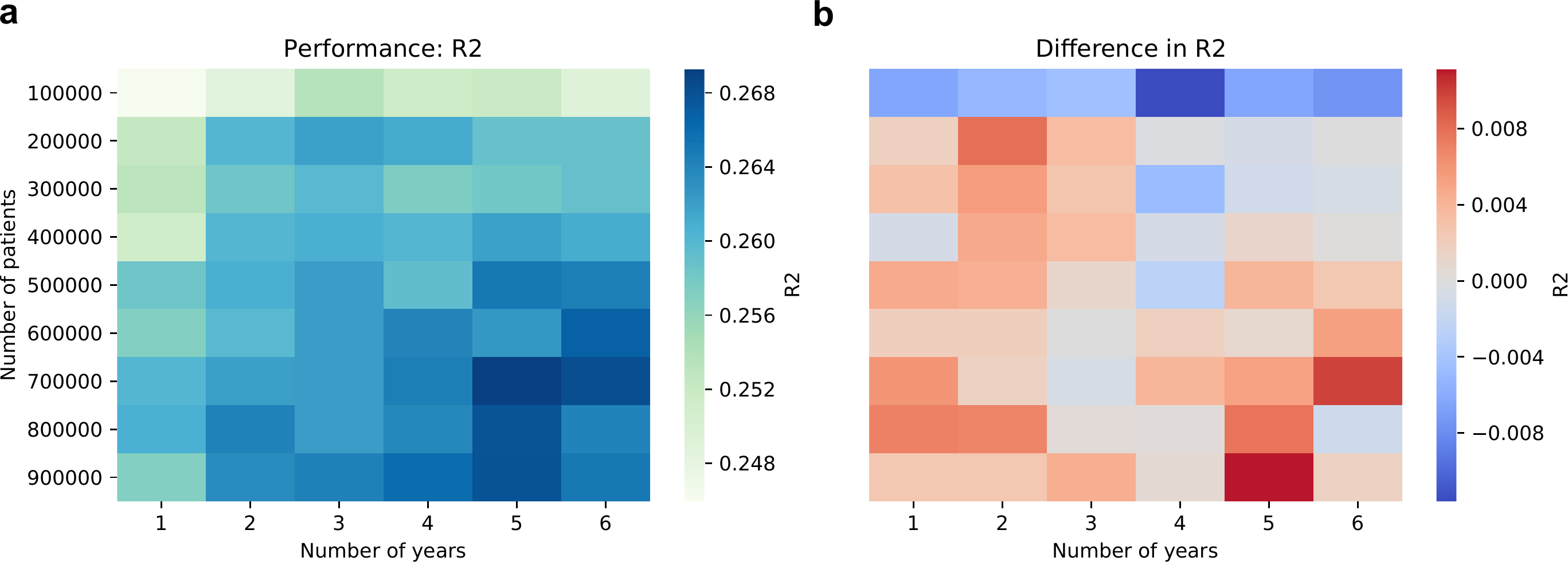}
	\end{center}
	\caption{\textbf{Dependence of performance on patient number and observation time:} Shown is the performance ($r^2$) of the neural network depending on the patient number and the length of the observation period in years \textbf{(a)}. Shown in \textbf{(b)} is the difference between the $r2$ of the neural network and of the ridge regression.}
	\label{Fig:HeatMap:R2:Diff_1}
\end{figure}

As sensitivity analyses, we studied how the number of samples in the training set and the length of the observation period affect the performance of the prediction for the neural network. Our analyses showed that as the number of patients increased the predictive performance , as measured by $R^2$, increased.  The same was true when the observation time increased (See Figure~\ref{Fig:HeatMap:R2:Diff_1}a). A similar picture can also be seen for the Spearman and Pearson correlation (See Supplementary Figure~S\ref{SupFig:Perf:nn}).
We compared this to the performance of the ridge regression (See Figure~\ref{Fig:HeatMap:R2:Diff_1}b). We found that at $100'000$ patients the $r^2$ of the neural network was lower than for the ridge regression but that for larger sample sizes the neural network had in general a higher $r^2$.

\begin{figure}
	\begin{center}
		\includegraphics[width=\textwidth]{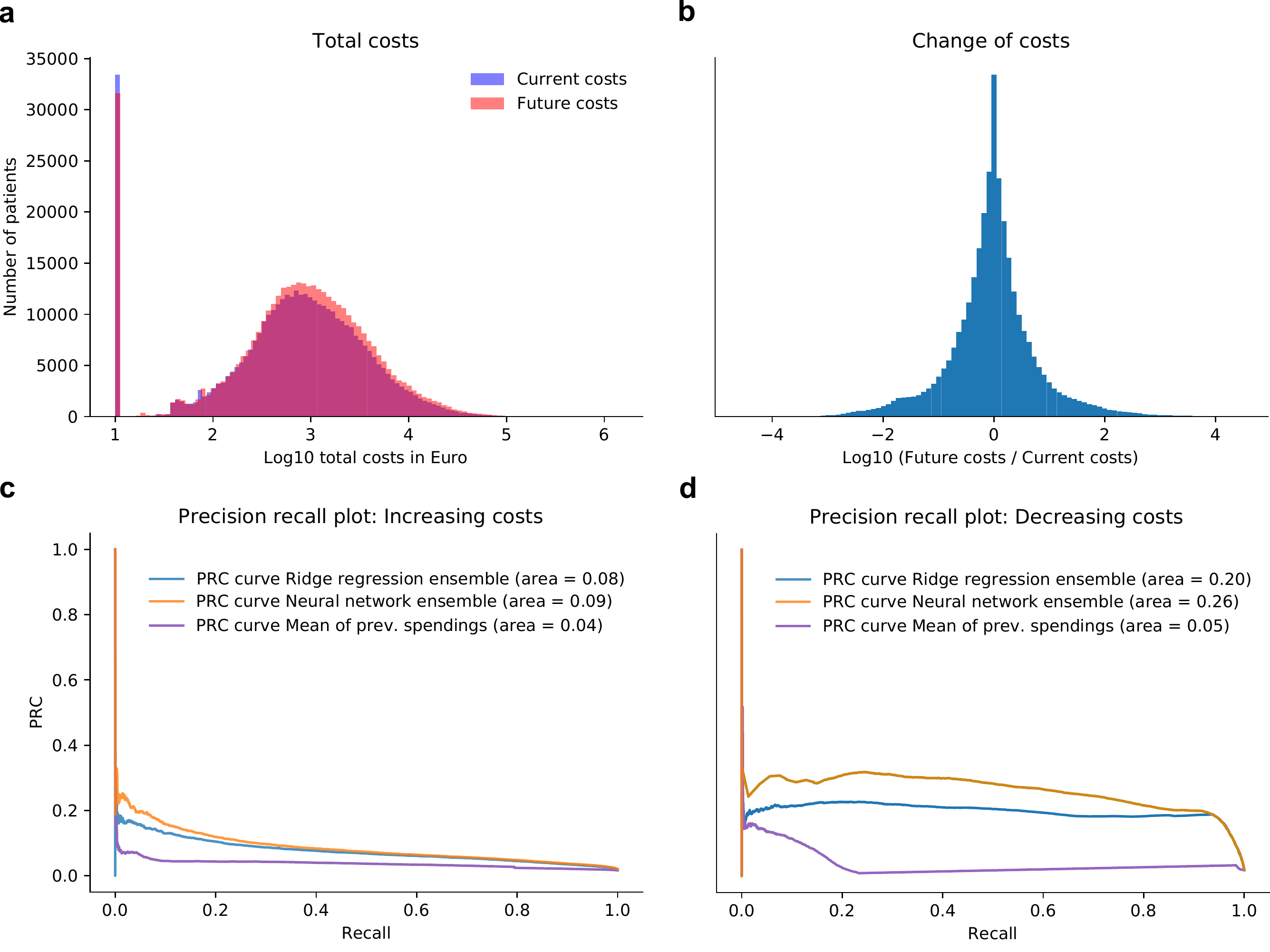}
	\end{center}
	\caption{\textbf{Cost change prediction:} Shown are the raw cost \textbf{(a)} in the last year of the observation period (current costs) and the evaluation period (Future costs) as well as the log10 fold-change between them \textbf{(b)}. Shown in \textbf{(c)}-\textbf{(d)} are the precision recall curves for predicting increasing and decreasing costs. }
	\label{Fig:Pref:TotalCost}
\end{figure}

Next, we analysed the ability of identifying patients with changing costs (Figure~\ref{Fig:Pref:TotalCost}a and Figure~\ref{Fig:Pref:TotalCost}b). In this analysis, we did not consider the model that used the last years costs as prediction for the future costs as costs are predicted to stay constant for this model. The results of the analysis in predicting patients with increasing/decreasing costs are shown in Figure~\ref{Fig:Pref:TotalCost}c and Figure~\ref{Fig:Pref:TotalCost}d, respectively. We found that overall, prediction of decreasing costs was easier than increasing costs. Furthermore, we found that for both direction of the cost change the neural network outperformed the ridge regression. For increasing costs the neural network had an auPRC of $0.08$ while the ridge regression only had an auPRC of $0.04$. For decreasing costs the neural network at an auPRC of $0.24$ while the ridge regression had an auPRC of $0.21$. 
A similar picture also emerged for the area under the ROC curve where the neural network had an auROC of $0.93$ and $0.90$ for decreasing and increasing costs, respectively. Here the ridge regression had an auROC of $0.93$ and $0.86$ for decreasing and increasing costs, respectively. 
For both measures the Ridge regression and the neural network were substantially better than the baseline methods that did not model the costs. 

\begin{figure}[h]
	\begin{center}
	\includegraphics[width=0.8\textwidth]{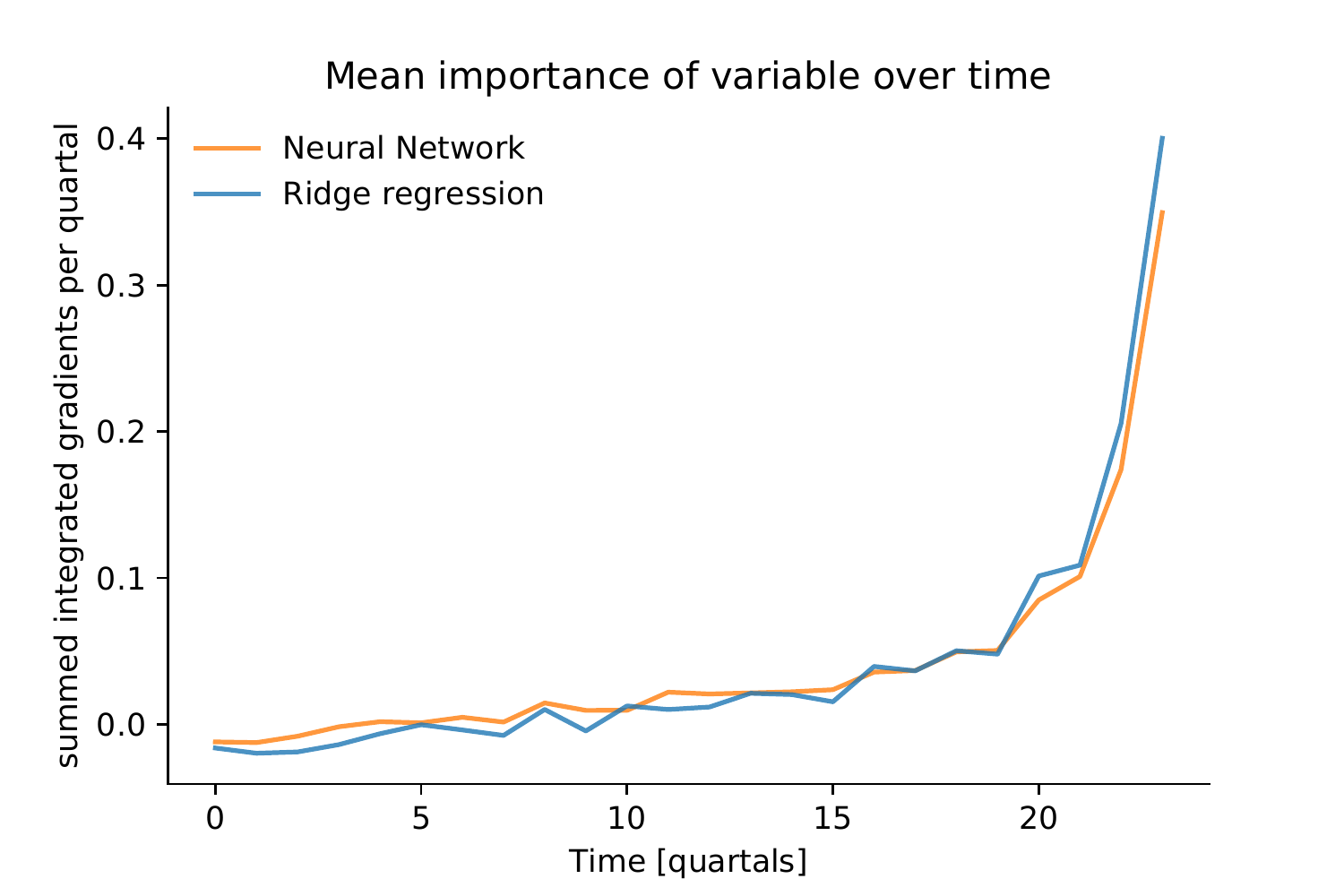}
	\end{center}
	\caption{\textbf{Importance of features per quarter:} Shown is the summed normalized integrated gradient (Importance) per year for the ridge regression and the neural network.}
	\label{Fig:TempImport}
\end{figure}

Finally, we studied via integrated gradients, on which features of the data the neural network based its prediction and how this differed from the features used by the ridge regression. 
We first determined the importance of features from different quarters in the observation period by summing the integrated gradients of all feature in a quarter. We found that both methods have a similar temporal distribution of the importance and that for prediction the most recent features were the most important (See Figure~\ref{Fig:TempImport}).

To evaluate whether the features showed a qualitative difference between the neural network and the regression, we identified the features with the highest (associated with higher cost) integrated gradient in set of patients that have an $100$-fold increase in costs.
To this end, we summed the integrated gradients of each code over all quarters in the observation period. The top-20 codes are shown in Table~\ref{Tab:IntGrdFeatures:NN:Up} for the neural network and for the ridge regression in Table~\ref{Tab:IntGrdFeatures:RR:Up}. We found that the neural network relied more on ICD10 diagnosis and ATC medication code than the ridge regression (8 of 20 vs. 3 of 20). 

\npdecimalsign{.}
\nprounddigits{3}
\begin{table}[]
	\begin{tabular}{p{3.2cm} p{6.5cm} N{3}{2}}\toprule
		Code & Short description & IG\\\midrule
		Sex Male & Male patient& 0.01663005677978025\\
		GOP key 01321 & Treatment by non health insurance-accredited physicians & 0.014202659114965158\\
		GOP key 03004 &  Practitioner care of patients with age between 55 and 75 years & 0.013801358033147486\\
		GOP key 03220 & Treatment  of a patient with at least one life-changing chronic disease & 0.011976154700583918\\
		ICD10 key F102 & Alcohol dependence & 0.011275252604278153\\
		Sex Female & Female patient & 0.010864753223592694\\
		GOP key 03221 &  Treatment  of a patient with at least one life-changing chronic disease & 0.010416274410177116\\
		GOP key 03005 &  Practitioner care of patients older than 75 years & 0.009679312657318166\\
		ICD10 key F171 & Mental and behavioural disorders due to use of tobacco & 0.009495032983072067\\
		ICD10 key M171 & Unilateral primary osteoarthritis of knee & 0.00882045299236326\\
		ATC code B01AC04 & Clopidogrel (antithrombotic agent) & 0.00864908326840559\\
		GOP key 03004R & Practitioner care of patients with age between 55 and 75 years  & 0.008617113271046145\\
		GOP key 32001A & Profitability bonus for arranging and providing laboratory services & 0.007746288146077733\\
		GOP key 01731 & Early detection of cancer in men & 0.007727717954102408\\
		ICD10 key M4809 & Spinal stenosis & 0.007686957182016932\\
		ICD10 key G359 & Multiple sclerosis & 0.007620857066320905\\
		OPS key 80103 & Application of drugs and electrolyte solutions via the vascular system in newborns & 0.00743787516312553\\
		ATC code N06AX16 & Venlafaxine (antidepressants) & 0.007160244804510887\\
		GOP key 32001B &  Profitability bonus for arranging and providing laboratory services & 0.007154926087990027\\
		GOP key 07220 & Surgical primary care  & 0.007033473779696349\\
		GOP key 06222 & Basic ophthalmic care & 0.007000174901559455\\
		ICD10 key M545 & Low back pain & 0.006741140154194145\\
		ICD10 key F430 & Acute stress reaction & 0.006637828850156036\\
	\end{tabular}
	\caption{Codes with the highest feature importance as determined by the integrated gradients (IG) for the neural network, as well as a description of the codes}
	\label{Tab:IntGrdFeatures:NN:Up}
\end{table}

\begin{table}[]
	\begin{tabular}{p{4cm} p{8cm} N{5}{2}}\toprule
			Code & Short description & IG\\\midrule
		GOP key 03220 & Treatment  of a patient with at least one life-changing chronic disease & 0.020124222053605728\\
		GOP key 03221 &  Treatment  of a patient with at least one life-changing chronic disease& 0.01669040130700444\\
		Sex Male & Male  patient & 0.016259663763891635\\
		Sex Female & Female patient & 0.01476920396855111\\
		GOP key 03004R & Practitioner care of patients with age between 55 and 75 years & 0.013557530518322416\\
		GOP key 03004 & Practitioner care of patients with age between 55 and 75 years & 0.012715576503986727\\
		GOP key 03040 & Family medical care & 0.012632908651367132\\
		GOP key 03230 & Problem-oriented medical consultation, which is necessary due to the nature and severity of the illness & 0.010322750311984236\\
		ATC code V04CA02 &  Tests for diabetes (Glucose) & 0.009901732570387062\\
		GOP key 03111R & Practitioner care of patients with age between 5 and 59 years & 0.007426489923709521\\
		ATC code M01AE01 & Ibuprofen (antiinflammatory and antirheumatic)& 0.007161158610574709\\
		ICD10 key F171 & Mental and behavioural disorders due to use of tobacco & 0.006974808563150814\\
		Hospital type & Hospitals of maximum care & 0.00649756895783582\\
		GOP key 32001J & Profitability bonus for arranging and providing laboratory services & 0.005921266294627465\\
		GOP key 03040G & General practitioner medical care & 0.005821182046680033\\
		GOP key 32001A & Profitability bonus for arranging and providing laboratory services & 0.0055376528611210335\\
		FG key 53 &  Physician group: Neurology & 0.0054575652119783725\\
		GOP key 32094 &  Quantitative determination of glycated hemoglobins& 0.005415745678649049\\
		GOP key 03212 & Chronic disease & 0.005138682908537243\\
		FG key 11 & Physician group: Trauma Surgery & 0.005012835700628195\\
		GOP key 01321 & Treatment by non health insurance-accredited physicians & 0.004978426230296797\\
	\end{tabular}
	\caption{Codes with the highest feature importance as determined by the integrated gradients (IG) for the ridge regression, as well as a description of the codes}
	\label{Tab:IntGrdFeatures:RR:Up}
\end{table}

\section{Discussion}
Accurate prediction of future health care cost provides the basis to optimally manage healthcare costs. Furthermore, identification of patients whose cost will change allows optimization of interventions given a limited budget in order to improve population health.	
To achieve this it is important to have accurate predictions of the future health costs. In this work we presented a deep learning based approach to predict future costs. Our approach can leverage the full complexity of the patient records and does not require prior feature selection.

We showed that our approach can outperform standard approaches, including the Morbi-RSA for all measured performance metrics (See Tab.\ref{Tab:Perfm}). We suggest that the performance gain is due to two reasons. First, our approach learns important features from the data and does not require manual feature selection. It has been shown that learnt features allow better predictions in computer vision and speech processing given enough training data~\cite{lecun2015deep}).
The value of learning predictive features from the data is suggested by the better (state-of-the-art) performance of our implementation of ridge regression compared to the existing implementation of Morbi-RSA that is only based on $80$ diseases.
Second, our deep learning approach allows modelling of complex interactions between all variables which is not possible for ridge regression. This enables better modelling of medical phenotypes such as interactions between age, sex and diagnosis. This is supported by the identified terms that are associated with increasing costs between ridge regression and the deep neural network, where the ridge regression uses mainly the GOP codes and the deep network puts a higher emphasis on medical diagnoses and prescribed drugs. It also worth noting that in contrast to the Morbi-RSA, which is mainly based on ICD10 codes, both the ridge regression and the neural network rely on GOP codes.
 
Since we placed no strong assumption on the phenotype that we modelled, we believe that the neural network may also easily adapted to predict other medical phenotypes.
However, we also acknowledge that further research is necessary to better understand the merits and limits of deep learning in identifying medical phenotypes from insurance claims. This includes the optimal architecture of the networks but also strategies to interpret deep networks, to provide uncertainty estimates for the models and model distribution shifts caused by changes in billing and treatment guidelines. 

\section{Conflicts of interest/Competing interests} None declared.

\section{Availability of data and material} The datasets used in the current study are not publicly available due to privacy and security concerns.

\section{Code availability} Code will be available upon publication.

\section{Author contributions} 
Conceptualization: Philipp Drewe-Boss, Jochen Walker, Uwe Ohler; Methodology: Philipp Drewe-Boss; Formal analysis and investigation: Philipp Drewe-Boss; Writing - original draft preparation: Philipp Drewe-Boss; Writing - review and editing: Philipp Drewe-Boss, Dirk Enders, Jochen Walker, Uwe Ohler; Funding: Uwe Ohler, Jochen Walker, Philipp Drewe-Boss.

\section{Acknowledgements}

The authors would like to thank Tina Ploner, Wolfgang Galetzka, Thomas M\"{u}hlenhoff and Wolfgang Kopp for their input. The authors would furthermore thank NVIDIA Corporation for the donation of a Titan Xp GPU used for this research. Finally, the authors would like to thank Philipp Jordan for help with the graphical abstract.

\section{Bibliography}
 
\bibliographystyle{plainnat}
\bibliography{bibliography}

\begin{thebibliography}{19}
\providecommand{\natexlab}[1]{#1}
\providecommand{\url}[1]{\texttt{#1}}
\expandafter\ifx\csname urlstyle\endcsname\relax
  \providecommand{\doi}[1]{doi: #1}\else
  \providecommand{\doi}{doi: \begingroup \urlstyle{rm}\Url}\fi

\bibitem[Andersohn and Walker(2015)]{Andersohn2015}
Frank Andersohn and Jochen Walker.
\newblock Characteristics and external validity of the german health risk
  institute ({HRI}) database.
\newblock \emph{Pharmacoepidemiology and Drug Safety}, 25\penalty0
  (1):\penalty0 106--109, nov 2015.
\newblock \doi{10.1002/pds.3895}.

\bibitem[Bertsimas et~al.(2008)Bertsimas, Bjarnad{\'{o}}ttir, Kane, Kryder,
  Pandey, Vempala, and Wang]{Bertsimas2008}
Dimitris Bertsimas, Margr{\'{e}}t~V. Bjarnad{\'{o}}ttir, Michael~A. Kane,
  J.~Christian Kryder, Rudra Pandey, Santosh Vempala, and Grant Wang.
\newblock Algorithmic prediction of health-care costs.
\newblock \emph{Operations Research}, 56\penalty0 (6):\penalty0 1382--1392, dec
  2008.
\newblock \doi{10.1287/opre.1080.0619}.

\bibitem[Chollet et~al.(2015)]{chollet2015keras}
Fran\c{c}ois Chollet et~al.
\newblock Keras.
\newblock \url{https://github.com/fchollet/keras}, 2015.

\bibitem[Diehr et~al.(1999)Diehr, Yanez, Ash, Hornbrook, and Lin]{Diehr1999}
P.~Diehr, D.~Yanez, A.~Ash, M.~Hornbrook, and D.~Y. Lin.
\newblock {METHODS} {FOR} {ANALYZING} {HEALTH} {CARE} {UTILIZATION} {AND}
  {COSTS}.
\newblock \emph{Annual Review of Public Health}, 20\penalty0 (1):\penalty0
  125--144, may 1999.
\newblock \doi{10.1146/annurev.publhealth.20.1.125}.

\bibitem[Dr{\"o}sler et~al.(2017)Dr{\"o}sler, Garbe, Hasford, Schubert, Ulrich,
  van~de Ven, Wambach, Wasem, and Wille]{drosler2017sondergutachten}
Saskia Dr{\"o}sler, Edeltraud Garbe, J~Hasford, I~Schubert, V~Ulrich, WPMM
  van~de Ven, A~Wambach, J~Wasem, and E~Wille.
\newblock Sondergutachten zu den wirkungen des morbidit{\"a}tsorientierten
  risikostrukturausgleichs.
\newblock \emph{Bonn, Wissenschaftlicher Beirat zur Weiterentwicklung des
  Risikostrukturausgleichs beim Bundesversicherungsamt im Auftrag des
  Bundesministeriums f{\"u}r Gesundheit}, 2017.

\bibitem[Dr{\"o}sler et~al.(2018)Dr{\"o}sler, Garbe, Hasford, Schubert, Ulrich,
  van~de Ven, Wambach, Wasem, and Wille]{drosler2018}
Saskia Dr{\"o}sler, Edeltraud Garbe, J~Hasford, I~Schubert, V~Ulrich, WPMM
  van~de Ven, A~Wambach, J~Wasem, and E~Wille.
\newblock Gutachten zu den regionalen verteilungswirkungen des
  morbidit\"{a}tsorientierten risikostrukturausgleichs.
\newblock \emph{Bonn, Wissenschaftlicher Beirat zur Weiterentwicklung des
  Risikostrukturausgleichs beim Bundesversicherungsamt im Auftrag des
  Bundesministeriums f{\"u}r Gesundheit}, 2018.

\bibitem[Duncan et~al.(2016)Duncan, Loginov, and Ludkovski]{Duncan2016}
I.~Duncan, M.~Loginov, and M.~Ludkovski.
\newblock Testing alternative regression frameworks for predictive modeling of
  health care costs.
\newblock \emph{North American Actuarial Journal}, 20\penalty0 (1):\penalty0
  65--87, jan 2016.
\newblock \doi{10.1080/10920277.2015.1110491}.

\bibitem[Esteva et~al.(2017)Esteva, Kuprel, Novoa, Ko, Swetter, Blau, and
  Thrun]{esteva2017dermatologist}
Andre Esteva, Brett Kuprel, Roberto~A Novoa, Justin Ko, Susan~M Swetter,
  Helen~M Blau, and Sebastian Thrun.
\newblock Dermatologist-level classification of skin cancer with deep neural
  networks.
\newblock \emph{Nature}, 542\penalty0 (7639):\penalty0 115, 2017.

\bibitem[Frees et~al.(2013)Frees, Jin, and Lin]{Frees2013}
Edward~W. Frees, Xiaoli Jin, and Xiao Lin.
\newblock Actuarial applications of multivariate two-part regression models.
\newblock \emph{Annals of Actuarial Science}, 7\penalty0 (2):\penalty0
  258--287, apr 2013.
\newblock \doi{10.1017/s1748499512000346}.

\bibitem[Gulshan et~al.(2016)Gulshan, Peng, Coram, Stumpe, Wu, Narayanaswamy,
  Venugopalan, Widner, Madams, Cuadros, et~al.]{gulshan2016development}
Varun Gulshan, Lily Peng, Marc Coram, Martin~C Stumpe, Derek Wu, Arunachalam
  Narayanaswamy, Subhashini Venugopalan, Kasumi Widner, Tom Madams, Jorge
  Cuadros, et~al.
\newblock Development and validation of a deep learning algorithm for detection
  of diabetic retinopathy in retinal fundus photographs.
\newblock \emph{Jama}, 316\penalty0 (22):\penalty0 2402--2410, 2016.

\bibitem[Kingma and Ba(2014)]{kingma2014adam}
Diederik~P Kingma and Jimmy Ba.
\newblock Adam: A method for stochastic optimization.
\newblock \emph{arXiv preprint arXiv:1412.6980}, 2014.

\bibitem[Lahiri and Agarwal(2014)]{lahiri2014predicting}
Bibudh Lahiri and Nitin Agarwal.
\newblock Predicting healthcare expenditure increase for an individual from
  medicare data.
\newblock In \emph{Proceedings of the ACM SIGKDD Workshop on Health
  Informatics}, 2014.

\bibitem[LeCun et~al.(2015)LeCun, Bengio, and Hinton]{lecun2015deep}
Yann LeCun, Yoshua Bengio, and Geoffrey Hinton.
\newblock Deep learning.
\newblock \emph{nature}, 521\penalty0 (7553):\penalty0 436, 2015.

\bibitem[Nair and Hinton(2010)]{nair2010rectified}
Vinod Nair and Geoffrey~E Hinton.
\newblock Rectified linear units improve restricted boltzmann machines.
\newblock In \emph{Proceedings of the 27th international conference on machine
  learning (ICML-10)}, pages 807--814, 2010.

\bibitem[Rajkomar et~al.(2018)Rajkomar, Oren, Chen, Dai, Hajaj, Hardt, Liu,
  Liu, Marcus, Sun, et~al.]{rajkomar2018scalable}
Alvin Rajkomar, Eyal Oren, Kai Chen, Andrew~M Dai, Nissan Hajaj, Michaela
  Hardt, Peter~J Liu, Xiaobing Liu, Jake Marcus, Mimi Sun, et~al.
\newblock Scalable and accurate deep learning with electronic health records.
\newblock \emph{NPJ Digital Medicine}, 1\penalty0 (1):\penalty0 18, 2018.

\bibitem[Srivastava et~al.(2014)Srivastava, Hinton, Krizhevsky, Sutskever, and
  Salakhutdinov]{srivastava2014dropout}
Nitish Srivastava, Geoffrey Hinton, Alex Krizhevsky, Ilya Sutskever, and Ruslan
  Salakhutdinov.
\newblock Dropout: a simple way to prevent neural networks from overfitting.
\newblock \emph{The journal of machine learning research}, 15\penalty0
  (1):\penalty0 1929--1958, 2014.

\bibitem[Sundararajan et~al.(2017)Sundararajan, Taly, and
  Yan]{sundararajan2017axiomatic}
Mukund Sundararajan, Ankur Taly, and Qiqi Yan.
\newblock Axiomatic attribution for deep networks.
\newblock In \emph{Proceedings of the 34th International Conference on Machine
  Learning-Volume 70}, pages 3319--3328. JMLR. org, 2017.

\bibitem[Sushmita et~al.(2015)Sushmita, Newman, Marquardt, Ram, Prasad, Cock,
  and Teredesai]{Sushmita2015}
Shanu Sushmita, Stacey Newman, James Marquardt, Prabhu Ram, Viren Prasad,
  Martine~De Cock, and Ankur Teredesai.
\newblock Population cost prediction on public healthcare datasets.
\newblock In \emph{Proceedings of the 5th International Conference on Digital
  Health 2015 - DH 15}. {ACM} Press, 2015.
\newblock \doi{10.1145/2750511.2750521}.

\bibitem[Tang et~al.(2018)Tang, Tam, Cadrin-Ch{\^e}nevert, Guest, Chong,
  Barfett, Chepelev, Cairns, Mitchell, Cicero, et~al.]{tang2018canadian}
An~Tang, Roger Tam, Alexandre Cadrin-Ch{\^e}nevert, Will Guest, Jaron Chong,
  Joseph Barfett, Leonid Chepelev, Robyn Cairns, J~Ross Mitchell, Mark~D
  Cicero, et~al.
\newblock Canadian association of radiologists white paper on artificial
  intelligence in radiology.
\newblock \emph{Canadian Association of Radiologists Journal}, 69\penalty0
  (2):\penalty0 120--135, 2018.

\end{thebibliography}
\newpage
\section{Supplement}
\renewcommand{\figurename}{Supplementary Figure}
\setcounter{figure}{0} 

\begin{figure}[!htbp]
	\subfloat{\includegraphics[width=0.5\textwidth]{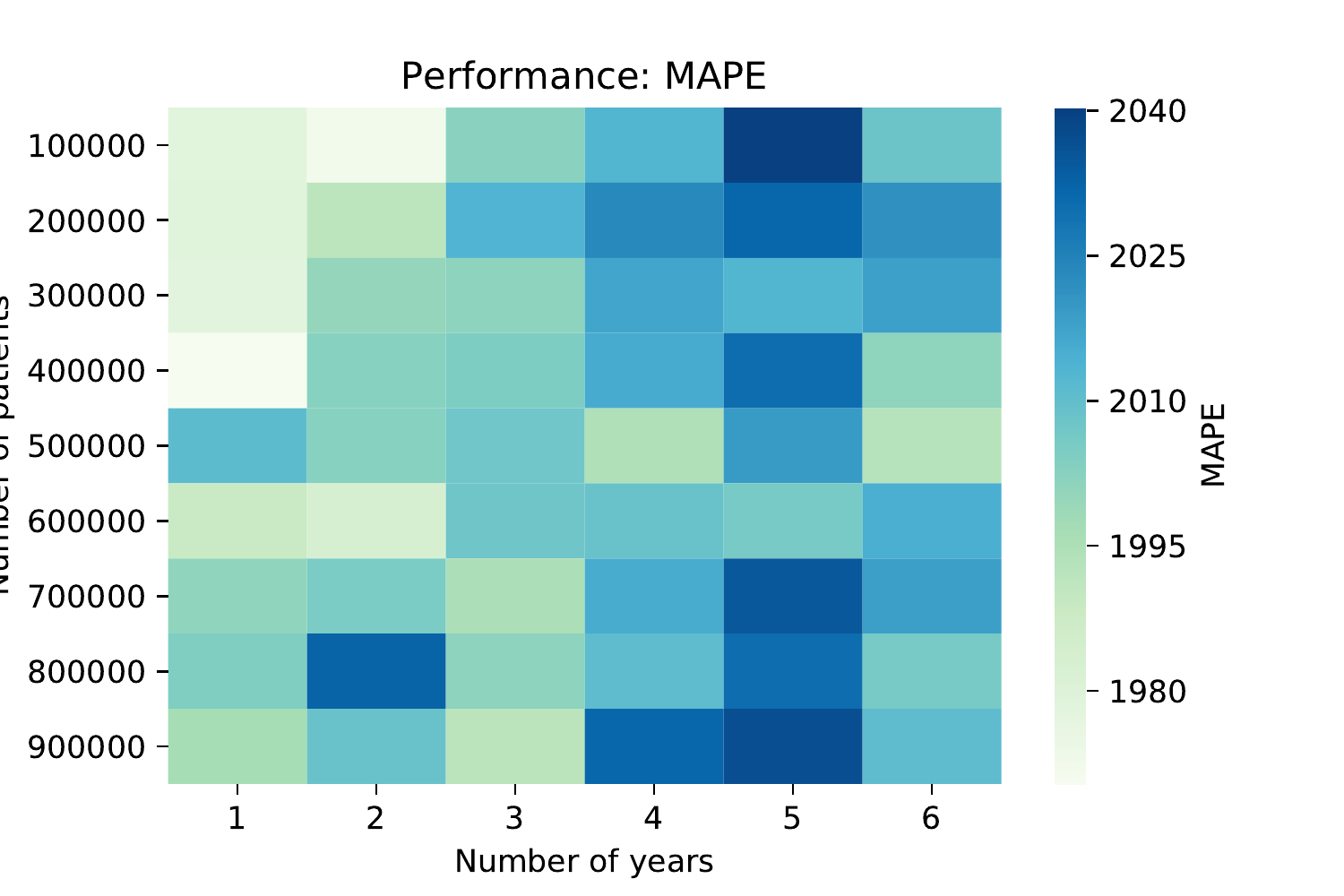}}%
	\subfloat{\includegraphics[width=0.5\textwidth]{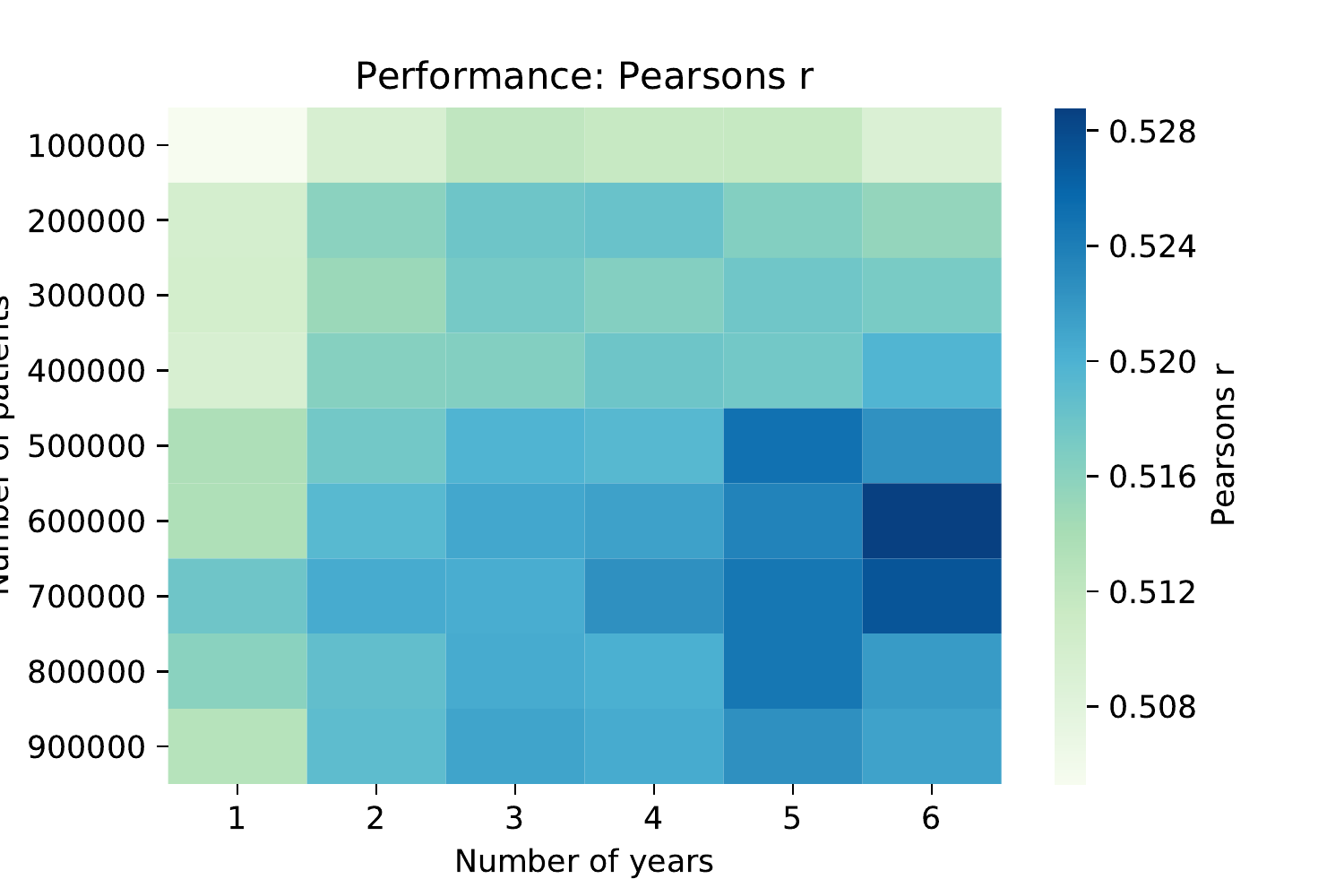}}%
	\hfill
	\subfloat{\includegraphics[width=0.5\textwidth]{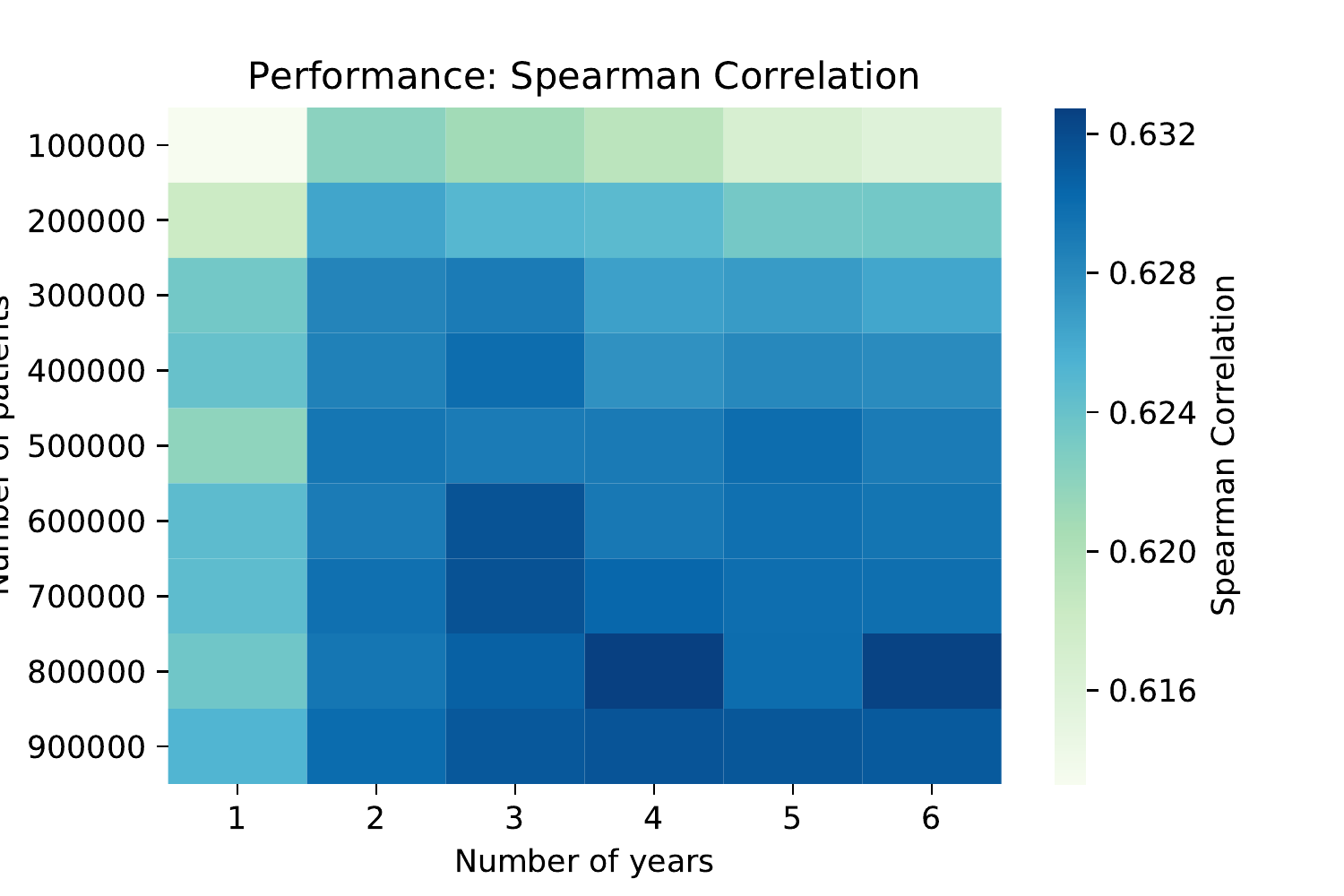}}%
	\subfloat{\includegraphics[width=0.5\textwidth]{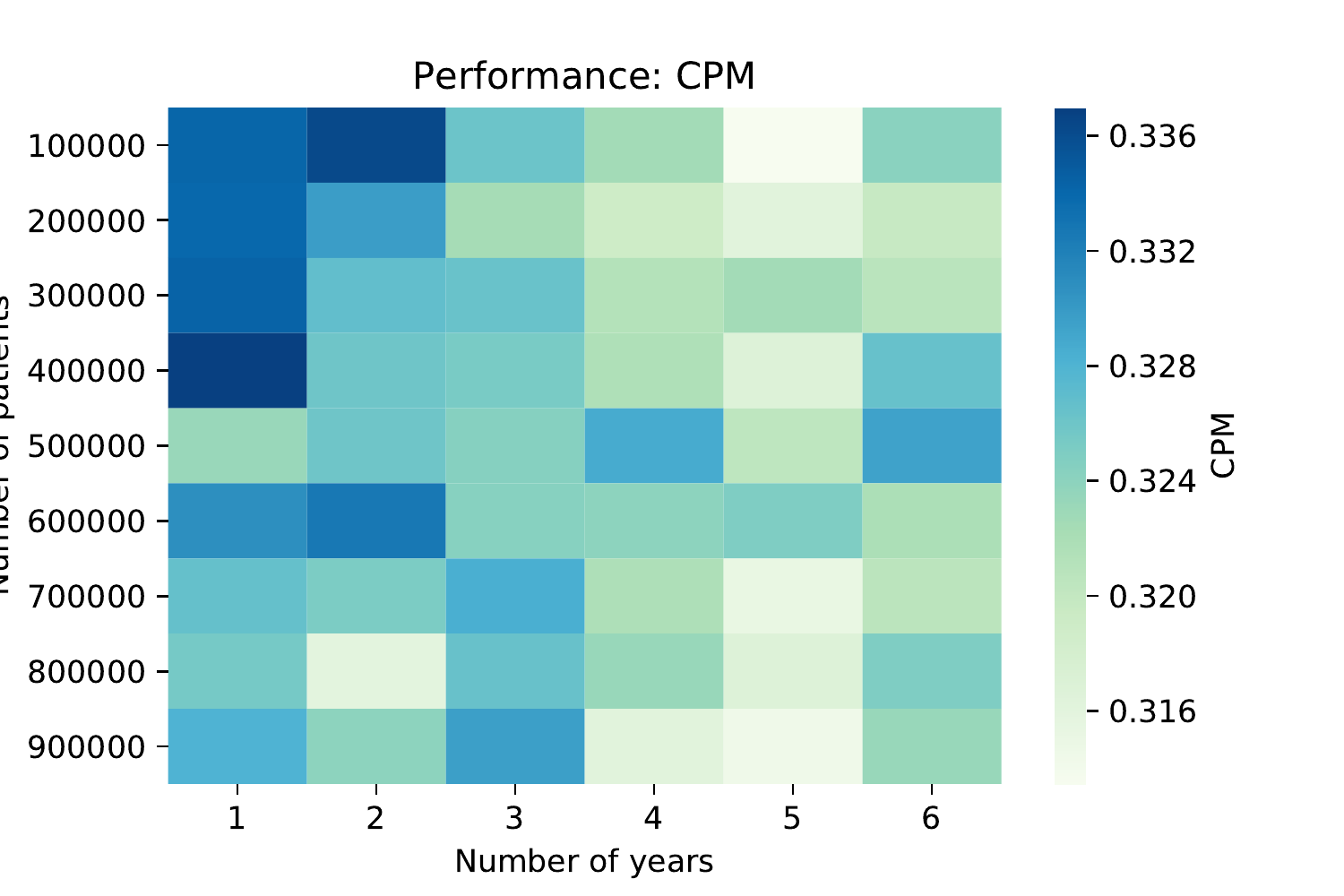}}
	\hfill
	\subfloat{\includegraphics[width=0.5\textwidth]{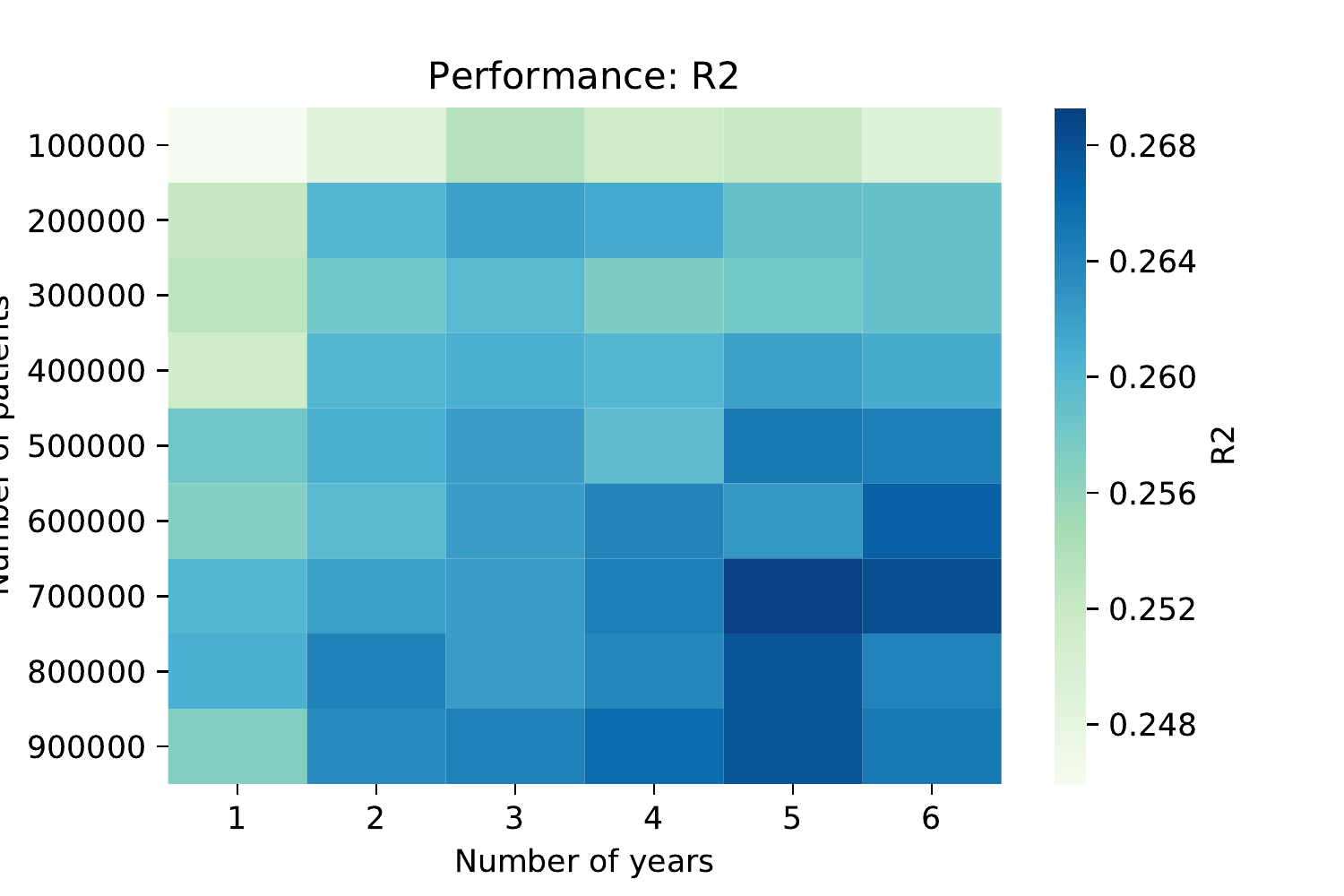}}
	\hfill
	\caption{\textbf{Dependence of performance on patient number and observation time:} Shown is the performance (Pearson's correlation (r),  Spearman's correlation ($\rho$),  mean absolute prediction error (MAPE), R squared ($r^2$) and Cumming' s Prediction Measure (CPM) of the neural network depending on the patient number and the length of the observation period in years.}
	\label{SupFig:Perf:nn}
\end{figure}

\end{document}